# Integrated Feature Selection for classification and Representation Learning with Group-Lasso Siamese Networks

Nisargo Ranjan Pal and Pradeep Singh, *Senior Member, IEEE*

***Abstract*—The Least Absolute Shrinkage and Selection Operator (Lasso) has been widely employed for feature selection in linear models and has also been extended to multi-layer perceptron (MLP) networks for both feature and group feature selection. Group-Lasso has also been used in feature selection using feed-forward neural networks. To the best of our knowledge, this is the first study to incorporate the Group-Lasso principle into a Siamese network architecture for feature selection. Unlike existing approaches that primarily optimize classification accuracy, the proposed method aims to identify a compact subset of features that simultaneously preserves the intrinsic representation of the input data (the goal of Siamese network) and is expected to achieve high classification performance. Two variants of the proposed framework are developed: one based on the Euclidean distance and the other on a modified Jaccard similarity measure for real-valued vectors. The latter is specifically designed to improve feature selection for high-dimensional sparse data. The effectiveness of the proposed approach is demonstrated on 12 benchmark datasets with feature dimensions ranging from 6 to 7070 using eight different classifiers. Experimental results demonstrate that the proposed Siamese Lasso framework consistently identifies compact and discriminative feature subsets maintaining competitive or superior classification performance, although the loss function used explicitly did not consider any regularizer to learn classification. In addition, the quality of the embeddings produced by these networks is assessed using manifold learning techniques such as UMAP and t-SNE.**

## I. Introduction

THE success of any clustering and classification algorithm significantly depends on the choice of features that represent an object[1]. Hence, the literature on feature analysis, which consists of feature selection and extraction, is very rich[2][3][4][5]. Feature selection can be done both with labelled and unlabelled data. When data are not labelled, the feature selection methods are called unsupervised. For labelled data, feature selection methods that exploit the class label are called supervised method. If the goal is to design a classifier, then a feature selection method tries to find useful features that can maximize the classification performance. In addition, a feature selection method should also try to eliminate derogatory features and control the use of dependent/redundant features [2]. For an unsupervised method, feature selection is more challenging as usually there is no target problem to solve. Thus, unsupervised methods try to optimize some task independent objective, such as preservation of substructure in the original data [4]. As a tool for feature selection, we can use statistical methods, neural networks, fuzzy systems, as well as evolutionary algorithms. Next, we briefly discuss some of the methods for feature selection.

Abdollahzadeh & Gharehchopogh (2022)[6] proposed three new Meta-Heuristic Algorithms for Feature Selection, the Harris Hawks Optimization (HHO) algorithm, the Fruitfly Optimization Algorithm (FOA) and their hybrid MOHHOFOA which uses Multi-objective optimization. The Proposed MOHHOFOA outperforms MOPSO and NSGA-II in classification accuracy and feature reduction even though the computational complexity increases with hybrid meta-heuristics. Gong et al. (2024)[7] proposed the CONMI_FS algorithm based on CONMI (Consistency and Mutual Information), which identifies the optimal subset of features that exhibit high correlation with the features while maintaining low redundancy among the selected features. Gong et. al (2024)[7]combined Pearson correlation (linear) and Mutual Information (non-linear) into a CONMI function and captured both linear and non-linear dependencies, improving diversity in feature evaluation but such pure filter methods may ignore feature interactions with specific classifiers. Nag et. al [8] proposed an algorithm that simultaneously selects useful features and designs diverse classifiers using a steady-state multi-objective genetic programming (GP). It optimizes three objectives: 1) false negatives (FNs); 2) false positives (FPs); and 3) the number of leaf nodes in the tree. The Framework can efficiently handle high feature-to-sample ratios and designs diverse ensembles at the cost of high computational expenses on high dimensional datasets.

Xiang et al. (2012)[9] introduced "ε-dragging" to enlarge distances between classes and used $L_{2,1}$-norm for sparsity. This "ε-dragging" concept is incorporated into Least Square Regression (LSR) problem for multi-class classification. This

Nisargo Ranjan Pal is with Kalinga Institute of Industrial Technology, Bhubaneswar, Odisha 751024, India
(email: 26167081@kiit.ac.in)

Pradeep Singh is with the department of Computer Sc & Engineering, National Institute of Technology Raipur 492010, India
( email: psingh.cs@nitrr.ac.in)

framework is called Discriminative LSR (DLSR). With its compact form, this model is extended for feature selection with the help of $L_{2,1}$ norm. An important feature of this method is that it does not require training independent two-class classifiers. Tao et al. (2016)[10] proposed a feature selection scheme that combines linear discriminant analysis (LDA) and sparsity regularization. To facilitate feature selection, this method induces row sparsity on the transformation matrix of LDA through $L_{2,1}$ regularization. This method can find discriminative features while discarding the redundant ones. Authors also considered the $L_{2,p}$-norm regularized variant that may yield better sparsity when $0 < p < 1$. Owing to this, [10] achieves better sparsity and discriminative power by selecting non-trivial solutions. However, non-convexity of $L_{2,p}$-norm (0<p<1) requires complex optimization. Yu & Liu (2003) [11] used "predominant correlation" to develop a filter-based method to identify relevant features along with removal of redundancy, which is highly efficient for high-dimensional data as it avoids pairwise correlation analysis but may miss complex non-linear feature interactions.

Gui et al. (2017)[12] conducted a survey on various feature selection methods based on structured sparsity. They proposed a taxonomy to explain the evolution of various sparsity-based methods. Gui et al. [12] categorized methods into vector-based (Lasso) and matrix-based ($L_{p,1}$-norm) categories. The authors discussed the differences and commonalities among these methods from the perspectives of regression and regularization. They also provided practical guidelines for feature selection in related applications. However, authors did not propose a single "best" algorithm rather focused on the taxonomy. Hou et al. (2014)[13] employed Joint Embedding Learning (JELSR) for unsupervised framework, which combine embedding learning and sparse regression ($L_{2,1}$-norm). Unlike conventional unsupervised feature selection techniques, their method combines the strengths of sparse regression and embedding learning. Across various datasets—including biological, voice, and image data—they demonstrate encouraging experimental outcomes.

Niijima & Okuno (2009)[14] adapted Laplacian Linear Discriminant Analysis (LLDA) for unsupervised feature selection. They developed an efficient algorithm to compute LLDA for high-dimensional, small-sample datasets (such as microarrays), resulting in a new technique called LLDA-based Recursive Feature Elimination (LLDA-RFE). This method [14] outperformed Laplacian score and SVD-entropy in biomarker discovery but incurred a high computational cost when used with RFE on extremely large feature sets. Bai et al. (2019)[15] used graph-based feature representations and Fused Lasso, which integrates kernel-based structural feature graphs with Fused lasso regularization to jointly maximize feature relevance and minimize redundancy. Their experimental results on several benchmark datasets using 10-fold cross-validation, demonstrated the effectiveness of the method.

Sun *et al.* (2017) [16] introduced a two-stage iterative LASSO-MLP algorithm that alternates between MLP training and LASSO-based pruning of input weights, with the regularization parameter selected through cross-validation. The method repeatedly removes zero-weight variables until no further improvement in prediction accuracy is achieved, yielding a sparse MLP with fewer input variables. Experimental evaluation on benchmark simulations and a real-world crude distillation process show that the proposed approach consistently outperforms several existing feature selection techniques in terms of prediction accuracy and model compactness. Wang *et al.* (2020)[17] proposed a neural network-based embedded feature selection framework that combines Group Lasso regularization with a redundancy-control penalty to simultaneously promote group-wise sparsity and minimize feature dependence. Since Group Lasso is not differentiable at the origin, the optimization employs a smoothed $L_{2,1}$-norm regularization with proven convergence, enabling stable training despite the non-smooth objective. Extensive experiments on both synthetic and real-world datasets demonstrated that the proposed method consistently selected less redundant and more discriminative feature subsets, outperforming conventional Group Lasso-based neural network feature selection methods while maintaining competitive predictive performance.

In this paper, we propose a Siamese network that simultaneously performs discriminatory feature selection and embedding learning. A Siamese network, through contrastive loss, learns representations that capture the relationships between pairs of data points. Once trained, it can compare previously unseen sample pairs without retraining, making it effective for applications such as signature verification and one-shot learning. Rather than learning class-specific decision boundaries, it learns an embedding space in which similar samples are close and dissimilar samples are far apart.

Despite their success in learning informative embeddings, conventional Siamese networks do not perform explicit feature selection. We therefore investigate whether a suitably modified Siamese network can simultaneously identify a subset of informative features while learning the desired embedding space, and whether the selected features are also effective for classification. If so, the learned feature subset could benefit not only similarity assessment but also other machine learning models, such as classifiers.

Reducing the number of input features lowers computational and storage requirements while also reducing data acquisition costs. This is particularly important for non-image data, where each feature often corresponds to a physical measurement, laboratory test, or sensor reading. Since a Siamese network must identify the features that bring instances from the same class closer together and push instances from different classes farther apart, these features are expected to be highly discriminative for classification also. To achieve this, we incorporate a Group Lasso regularizer into the Siamese loss. We investigate two variants of the proposed network based on different dissimilarity measures. To the best of our knowledge, this is the first study to integrate Group Lasso into a Siamese network for discriminative feature selection while preserving its primary similarity-learning objective.

## II. Siamese Network

A Siamese neural network [18] is a type of deep learning architecture designed to determine the similarity between two inputs by passing them through two identical subnetworks ("twins") with shared weights and parameters. Instead of classifying an object directly, the network categories the input pair of data as similar pairs and dissimilar pairs where similar items are close together and dissimilar items are far apart. The network outputs a *distance* or *similarity score*, to determine if two inputs are from the same class/similar or from different class/dissimilar. Usually, networks like MLP and CNN are used as the constituent identical sub-networks. Specialized loss functions like *contrastive loss* are used to minimize distance between similar pairs and maximize distance between dissimilar ones in the embedded data. Siamese networks are particularly effective for tasks with limited data per class, known as one-shot learning. Applications of Siamese network include comparing signatures in a banking system, identifying if two photos belong to the same person, detecting forgery or matching similar items. Sometimes it is also used to determine if two sentences or pieces of text carry the same meaning.

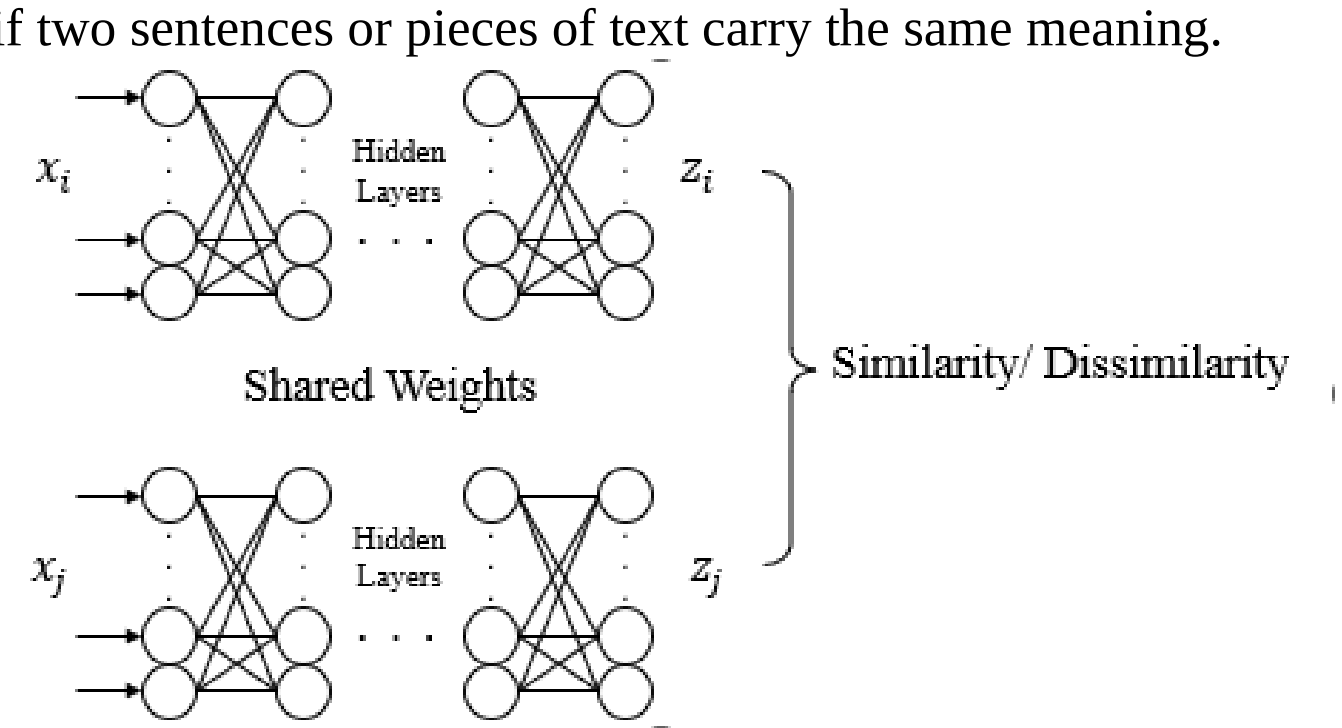

Fig 1. Siamese Network

### *2.1 Architecture*

Fig.1 shows the architecture of a Siamese network. It consists of two identical subnetworks or towers. In Fig. 1, $\boldsymbol{x}_i \in R^d$ and $\boldsymbol{x}_j \in R^d$ are inputs for the first and second sub-networks respectively, where $d$ is the dimensionality of the inputs, i.e., the number of input nodes in each of the subnets. Together, they form an input pair i.e., $\{\boldsymbol{x_i}, \boldsymbol{x_j}\}; \forall i, j; i, j = 1, \dots, n,$ is an input pair that is used for training the network. Likewise, $\boldsymbol{z}_i \in R^q$ and $\boldsymbol{z}_j \in R^q; q < d$ are the embeddings formed by the first and second sub-networks respectively. The dimension, $q$, of the embeddings depends on the training data set and the problem being solved, which should be defined by the user. Both sub-networks have the same number of hidden layers and they share the same weights. The number of hidden layers, like any feedforward network, such as multilayer perceptrons (MLPs), is a hyperparameter and its value depends on the data set used to train the system.

### *2.2 Processing and Embeddings*

To begin with both networks are randomly initialized using the same weights. Then input pairs like $\{\boldsymbol{x}_i, \boldsymbol{x}_j\}$ is fed into the network. The inputs $\boldsymbol{x}_i$ and $\boldsymbol{x}_j$ pass through the first and second sub-networks respectively. Both sub-networks run parallelly and independently. Each sub-network produces a low-dimensional feature vector called embedding that represent the input. Here, $\boldsymbol{z}_i$and $\boldsymbol{z}_j$ are the respective embeddings produced by the twin networks corresponding to a pair of inputs $\boldsymbol{x}_i$ and $\boldsymbol{x}_j$ . These embeddings capture the essential characteristics of the input such that similar pairs are closer and dissimilar pairs are apart.

Typically, a distance metric *D,* for example, Euclidean Distance ($L_2$ Norm) is used to calculate the distance between $\boldsymbol{z}_i$ and $\boldsymbol{z}_j$. This determines the similarity/ dissimilarity between the pair:

- If $\boldsymbol{x}_i$ and $\boldsymbol{x}_j$ are similar or are from the same class, then $d(\boldsymbol{z}_i, \boldsymbol{z}_j)$ should be low
- If $\boldsymbol{x}_i$ and $\boldsymbol{x}_j$ are dissimilar or coming from different classes, then $d(\boldsymbol{z}_i, \boldsymbol{z}_j)$ should be high.

### *2.3 Training and Loss Functions*

The network is then trained using a specialized loss function, called contrastive loss, as explained next, that aid in separation of similar and dissimilar input pairs. The Contrastive Loss tries minimizes the distance for matching (or similar) pairs and maximize the distance for differing/dissimilar pairs. For example, if $\boldsymbol{x_i}$ and $\boldsymbol{x_j}$ belong to the same class, then they are considered similar and the distance between the corresponding pair of embeddings $\boldsymbol{z_i}$ and $\boldsymbol{z_j}$ should be low**,** otherwise, it should be high. To define the contrastive loss, let us define a variable $y_{ij}$ for a pair $\{\boldsymbol{x_i}, \boldsymbol{x_j}\}$ such that $y_{ij} = 0$ when $\boldsymbol{x_i}$ and $\boldsymbol{x_j}$ are similar and $y_{ij} = 1$ **,** when $\boldsymbol{x_i}$ and $\boldsymbol{x_j}$ are dissimilar. Then the contrastive loss function is defined as:

$$L_{Siamese}(\boldsymbol{W}; \boldsymbol{x_i}, \boldsymbol{x}_j) = \frac{1}{2}[(1 - y_{ij})D_{ij}^2 + y_{ij}(max\{0, m - D_{ij}^2\})] \quad (1)$$

where $D_{ij}$ is a measure of dissimilarity between $\boldsymbol{z_i}$ and $\boldsymbol{z_j}$ **.** For example, $D_{ij}$ could be the Euclidean distance or (1-cosine similarity) between $\boldsymbol{z_i}$ and $\boldsymbol{z_j}$. In (1), *m* is the margin that define a threshold for the dissimilarity. $\boldsymbol{W}$ is the weight vector consisting of all weights. The second term of the loss function penalizes the loss, if the square of the dissimilarity is lower than the margin and this term is active only when $y_{ij}$ = 1. Thus, for a dissimilar pair, the first term is zero and the second term also vanishes if the pair of embeddings are already adequately separated. On the other hand, when $y_{ij}$ = 0, i.e., when $\boldsymbol{x_i}$ and $\boldsymbol{x_j}$ are similar, the first term is reduced, i.e., $\boldsymbol{z_i}$ and $\boldsymbol{z_j}$ are brought closer**.** Both networks share the same $\boldsymbol{W}$. Thus, optimizing the objective in (1), the network extracts appropriate features to generate the embeddings such that similar pairs are closer and dissimilar pairs are apart in the projected space. Training the Siamese network involves learning the weight vector $\boldsymbol{W}$ by minimizing L in (1) usually via gradient descent or some of its variant. A Siamese network can also be trained using a triplet loss [19], which minimizes the distance between an anchor and a positive sample while maximizing the distance between the anchor and a negative sample. We restrict our study only to the contrastive loss.

## III. The Proposed Integrated Generation of Embeddings and Feature Selection

A Siamese network, by virtue of the contrastive loss, enables it to learn representations that capture the underlying relationships among data points. As explained earlier, once the network is trained, it can compare previously unseen pair of samples without needing any further retraining. Thus, this network is very effective for signature verification, one-shot learning, and tasks with limited labelled data. By learning a meaningful embedding space instead of class-specific decision boundaries, Siamese networks provide a framework for assessing whether a pair of imputes is similar or not. Although, the network can do a good job of feature extraction (embeddings), it does not/cannot do any explicit feature selection. Here we raise a question: can a Siamese network or a modified version of it do feature selection as well as feature extraction (via embedding)? If the answer is affirmative, then we shall get a smaller set of input features that can solve the Siamese network problem. Additionally, can such features be used for other purposes such as designing of classifiers? We argue that such a set of selected features although not intended for classifier design, will have the discriminating power required for classification. Our argument is based on the premise that if the selected features enable similar data points to be embedded closer together while separating dissimilar data points, then the resulting feature set should possess strong discriminative capability. Moreover, as discussed earlier, feature reduction can further improve the efficiency of a Siamese network by reducing computational requirements as well as the complexity and cost of model design and decision making. For example, in case of non-image data, reducing the number of features will reduce the time and cost for input generation.

To design a Siamese network that can do feature selection as well as generation of useful embeddings in an integrated manner, we make use of the Group-Lasso ($L_{2,1}$) regularization. Although group-lasso has been used for feature selection using MLPs [20][21][22], to the best of our knowledge, this is the first attempt to use group-lasso along with Siamese network. In order to identify a useless/derogatory feature, we need to set all weights connecting that feature to all nodes in the first hidden layer to zero or near zero. We try to achieve this using a penalty/regularizer using group lasso. Let $\boldsymbol{W}^{(k)} = \left(w_{1,k}^{(1)}, \dots, w_{N_1,k}^{(1)}\right)^T$ be the vector of weights connecting the k-th input to the $N_1$ nodes in the first hidden layer and $\boldsymbol{W} = (\boldsymbol{W}^{(1)}, \boldsymbol{W}^2, \dots, \boldsymbol{W}^d)$ be the matrix of all weight vectors connecting input layer to the first hidden layer. The $L_2$ norm of $\boldsymbol{W}^{(k)}$ is

$$L_2\left(\boldsymbol{W}^{(k)}\right) = \sqrt{\sum_{m=1}^{N_1} \left(W_{mk}^{(1)}\right)^2}; \; k = 1, 2, \dots, d. \quad (2)$$

Equation (2) defines the $L_2$ norm of the vector $\boldsymbol{W}^{(k)}$; *k=1,2, …, d*. If feature *k* is not relevant for the task at hand, we want $L_2\left(\boldsymbol{W}^{(k)}\right)$ to be zero or nearly zero.

Therefore, the group-lasso regularizer/penalty for feature selection is defined as:

$$L_{2,1}(\boldsymbol{W}) = \sum_{k=1}^{d} L_2\left(\boldsymbol{W}^{(k)}\right) \quad (3)$$

The group lasso penalty treats all weights connecting a given input feature to the neurons in the first hidden layer as one group. Thus, the group-lasso regularizer in (3) encourages all weights associated with an input feature to become zero together, thereby performing feature selection. Group lasso can reduce the weights of groups corresponding to redundant and derogatory features to almost or even completely zero, keeping the useful features as those are needed to minimize the contrastive loss. The severity of the regularizer (penalty) can be monitored by multiplying the regularizer in (3) by a non-negative constant, which we discuss later. Before writing the complete objective/loss function for the group-lasso Siamese network (GLSN), we want to explain two variants of the network that we shall use. Typically, the Euclidean distance is used as a measure of dissimilarity for the Siamese network. However, curse of dimensionality for high dimensional data, creates problem when using the Euclidean distance. To overcome this, we shall also use another dissimilarity measure based on Jaccard Similarity as explained next.

**Euclidean Distance:**

$$D_{Euclidean}(x_i, x_j) = \left\|x_i - x_j\right\|_2 \quad (4)$$

**Jaccard Similarity based distance:**

The Jaccard similarity (or Tanimoto Coefficient) is defined to compute the similarity between two sets or two binary vectors. It is computed as the ratio of cardinalities of the of the intersection and union of two sets. Ralaivola et. al [23][24] provided a definition of the Tanimoto Coefficient/Jaccard Similarity using dot products as follows:

$$S_{Jaccard}(\boldsymbol{x_i}, \boldsymbol{x_j}) = \frac{x_i . x_j}{\|x_i\|^2 + \|x_j\|^2 - x_i . x_j} = \frac{\sum_k x_{ik} x_{jk}}{\sum_k x_{ik}^2 + \sum_k x_{jk}^2 - \sum_k x_{ik} x_{jk}} \quad (5)$$

Equation (5) provides a measure of similarity that can be computed for any pair of vectors, real or binary. Tripp et. al [24] proved that

$$D_{Jaccard}(\boldsymbol{x_i}, \boldsymbol{x_j}) = \sqrt{1 - S_{Jaccard}(\boldsymbol{x_i}, \boldsymbol{x_j})} \quad (6)$$

is a valid distance metric. Thus, we can use it in place of Euclidean distance in Siamese network. It is well known that as the dimension increases the Euclidean distance starts losing it discriminating power. In fact, as dimensionality increases, the ratio of the distance to the closest point and that to the farthest point, approaches one[25]. However, in Jaccard similarity, the numerator indicates the alignment of the feature values between the two vectors, which is scaled down by the norms of vectors after subtracting the dot product of the vectors. Thus, even for high-dimensional data (which is usually sparse) distance based on Jaccard similarity is expected to do a better job.

Using the regularizer and a measure of dissimilarity we can define the modified loss function for the Siamese network as follows:

$$\frac{1}{\boldsymbol{M}} \sum_{\boldsymbol{ij}} \boldsymbol{L_{siamese}}(\boldsymbol{W}; \boldsymbol{x_i}, \boldsymbol{x_j}, \boldsymbol{y_{ij}}) + \lambda \sum_{k=1}^{d} L_2\left(\boldsymbol{W}^{(k)}\right) \quad (7)$$

where $L_{Siamese}(\boldsymbol{W}; \boldsymbol{x_i}, \boldsymbol{x_j}) = \frac{1}{2}[(1 - y_{ij}) D_{ij}^2 + y_{ij}(max\{0, m - D_{ij}^2\})]$.

In (7), $M$ is the number of pairs in a batch and $\lambda > 0$ is a hyperparameter that controls the severity of the penalty; a higher $\lambda$ will discard more features. We can have two choices of the dissimilarity measure, one using the Euclidean distance and other using the Jaccard similarity-based distance. These two choices of dissimilarity result in two variants of the Siamese network, we call then SNEU and SNJS. It is important to emphasize that the Siamese network's loss function (7) has the same objective of generating projections that can be used to compare a pair of inputs, but it does it using a smaller set of features. We have *not* used any regularizer to monitor the *discriminating* power of the selected features for classification because we wanted to preserve the primary objective of Siamese Network. We shall demonstrate that this smaller set of features also has good discriminating power for classification.

**Algorithm 1: Siamese Network Using Jaccard Distance Metric (SNJS)**

**Input: - Dataset pairs $\{(\boldsymbol{x_a}, \boldsymbol{x_b})_i\}_{i=1}^{N}$ with labels $\boldsymbol{y_i} \in \boldsymbol{0}, \boldsymbol{1}$, Regularization strength $\boldsymbol{\lambda}$, Margin $\boldsymbol{m}$, Learning rate $\boldsymbol{\eta}$, Initialized weights $\boldsymbol{\theta}$**

Output**: - Feature importance scores $\boldsymbol{\omega} \in \mathbb{R}^{\boldsymbol{d}}$, optimized embedding model $\boldsymbol{f_\theta}$**

**Initialize** network architecture $f_\theta$ based on input dimension $d$

**Set** $\boldsymbol{L_{min}} \leftarrow \infty$, patience_counter $\leftarrow 0$

**While** epoch < MaxEpochs **do**:

**For** each batch$\{(\boldsymbol{x}_a, x_b, y_{ab})\}$ **do**:

Generate embeddings $z_a = f_\theta(\boldsymbol{W}, \boldsymbol{x}_a), \boldsymbol{z}_b = f_\theta(\boldsymbol{x}_b)$

**Compute Jaccard Distance** $D_J$:

$$D_J = \frac{\sum(\boldsymbol{z}_a \cdot \boldsymbol{z}_b)}{\sum z_a^2 + \sum z_b^2 - \sum(z_a \cdot z_b)}$$

**Compute Contrastive Loss** $L_c$:

$$L_C = \frac{1}{N}\sum\left[\left(1 - y_{ij}\right)D_{ij}^2 + y_{ij}\left(max\left\{0, m - D_{ij}^2\right\}\right)\right]$$

**Compute Group-Sparsity Penalty** $L_{reg}$:

$$L_{reg} = \lambda \sum_{k=1}^{d} L_2\left(\boldsymbol{W}^{(k)}\right)$$

Compute Total Loss

$$L_{total} = L_c + L_{reg}$$

**Update Weights:**

$$\theta \leftarrow Adam(\nabla_\theta L_{total}, \eta)$$

**End For**

**If** $L_{total}$ plateaus **then** $\eta \leftarrow \eta \times 0.5$

**End While**

**For** each feature $k$ in $\{1, \dots, d\}$ do:

$$\omega_k = \left\|\boldsymbol{W}^{(k)}\right\|_2 = \sqrt{\sum_{j=1}^{h} W_{kj}^2}$$

**End For**

**Return** Ranked importance $\omega$ and $f_\theta$

We shall use $\omega_k$ to select/discard features because a lower value of $\omega_k$ indicates a that the k-th feature is not useful to achieve the objective of the Siamese network.

The algorithm for SNEU can be obtained by replacing the step for computing Jaccard Distance by Euclidean Distance. Rest of the algorithm remains the same.

## IV. Data Dets and Computational Protocols

### *4.1 Data Sets*

The data sets used in this study are summarized in Table 1. We have considered 12 benchmark data sets with widely different characteristics. The dimensionality of the data sets varies between 6 to 7070 and the data set size varies between 74 to 7000.

TABLE 1: DATASETS USED

| Dataset | No of Data Points | Dimensions | No of Classes |
|---|---|---|---|
| **Flea[1]** | 74 | 6 | 3 |
| **Seeds[2]** | 210 | 7 | 3 |
| **Pima[2]** | 768 | 8 | 2 |
| **Wine[2]** | 178 | 13 | 3 |
| **Vehicle[2]** | 846 | 18 | 4 |
| **Ionosphere[2]** | 351 | 33 | 2 |
| **Sonar[2]** | 208 | 60 | 2 |
| **ORL[3]** | 400 | 1024 | 10 |
| **WarpPie10P[3]** | 210 | 2420 | 10 |
| **Lung[4]** | 203 | 3312 | 5 |
| **Gisette[5]** | 7000 | 5000 | 2 |
| **Leukaemia[6]** | 72 | 7070 | 2 |

### *4.2 Computational Protocols*

In principle, given a data set, the network architecture can be selected using cross-validation. But this requires adequate data as well as high computational resources, particularly for high dimensional data. Since our objective is not to design an optimal architecture but to establish the validity of our concept, the network architecture is determined by an intuitive heuristic based on the input dimensionality $d$. For low-dimensional datasets, $d < 10$ , a single hidden layer is employed with $d + 10$ nodes, followed by an embedding layer of size $\frac{1}{2}d$. For medium-dimensional datasets $10 < d < 1000$, the architecture utilizes a single hidden layer where the node count is halved from the input to the embedding layer. For high-dimensional datasets $d \geq 1000$, a dual hidden layer structure is implemented. In this configuration, the node counts decrease linearly such that the difference between successive layers remains constant, satisfying the arithmetic progression:

$$d - N_1 = N_1 - N_2 = N_2 - q$$

[1] http://ggobi.org/book/data/flea.xml
[2] https://archive.ics.uci.edu/
[3] https://jundongl.github.io/scikit-feature/datasets.html
[4] https://lce.biohpc.swmed.edu/lungcancer/datasetsearch.php?datasetid=56
[5] https://www.csie.ntu.edu.tw/%7Ecjlin/libsvmtools/datasets/binary.html#gisette
[6] https://zenodo.org/records/2709491

where $N_k$ represents the $k^{th}$ hidden layer and $q$ represents the dimension of the embedding layer $\frac{1}{2}d$. This is summarized in Table 2.

TABLE 2. NETWORK ARCHITECTURE BASED ON DATASET DIMENSION

| Input Dimension (d) | No. of Hidden Layers | Hidden Layer Nodes | Embedding Dimension |
|---|---|---|---|
| $\boldsymbol{d < 10}$ | 1 | $d + 10$ | $\frac{1}{2}d$ |
| $\boldsymbol{10 < d < 1000}$ | 1 | $\frac{1}{2}d$ | $\frac{1}{4}d$ |
| $\boldsymbol{d \geq 1000}$ | 2 | Linear decay (Arithmetic Progression) | $\frac{1}{2}d$ |

For consistency of the experiments, a set of globally defined parameters are used or *all* data sets as given in Table 3.

TABLE 3. THE SET OF PARAMETERS USED FOR ALL DATA SETS

| Parameters | Value |
|---|---|
| **Initial Learning Rate** | 0.001 |
| **Lambda For Sparsity ($\boldsymbol{\lambda}$)** | 0.005 |
| **Epochs** | 10 |
| **Batch Size** | 64 |
| **Batches Per Epoch** | Default |

As for the "Initial Learning Rate", for the possibility of the loss reaching a plateau, we also used a learning rate reducer function: *ReduceLROnPlateau(monitor='loss', factor=0.5, patience=3, min_lr = 0.00001, verbose = 1)*. If three consecutive loss values are close, then this function reduces the learning rate by half and the minimum learning rate that can be allowed has been set to 0.00001.

However, for computational efficiency while using the Gisette Dataset, a few exceptions were made. For the Gisette Dataset the number of features is 5000 and number of instances is 7000, and thus the number of batches produced is very high. Because the data set is high dimensional and the number of batches is high, it demands huge computational overhead. To be able to complete the task with our limited resources, the "Batches Per Epoch (Default = 5980 batches for Gisette)" has been reduced to *half*, i.e., $\left(\frac{1}{2} * 5980 =\right) 2990$. And, to compensate for the unused pairs due to a smaller number of batches, the "Epochs" were doubled ( 10 * 2 = 20) for each network. Rest of the parameters remain the same.

## V. RESULTS

Two variants of the Siamese network are trained, one using Euclidean distance (SNEU) and other using the Jaccard Similarity (SNJS). After the network is trained, the features are ranked in ascender order of $L_2$ norm of the weight vectors corresponding to different features, i.e., with respect to $\omega_k$. Thus, we obtain two different rankings to select features from. When features are selected using the ordering produced by SNEU, we call it Euclidean distance-based feature selection (in sort EUFS); similarly, for SNJS we call it Jaccard similarity-based feature selection (in sort JSFS).

We evaluated the performance of the proposed feature selection approaches using eight classifiers, namely, Nearest Neighbour (NN)[26], Linear SVM[27], RBF SVM,[27] Random Forest (RF)[28], Decision Tree (DT)[29], AdaBoost[30], Multilayer Perceptron (MLP) network[31], and Naïve Bayes (NB)[32] on twelve benchmark datasets spanning a wide range of dimensionalities (from 6 features in Flea to 7070 features in Leukaemia). For each classifier, experiments are performed using both EUFS and JSFS. To check the performance of a classifier, we need to decide on the number features to use. Our goal is not to come up with an optimal set of features, but to show that while learning the task of Siamese network, it can also select a good set of discriminatory features for classification. So, we consider 5%, 10%, 15%, 20%, 25%, 30%, 35%, 40%, 45%, 50%, and 100% (all features). If we can demonstrate that with a small set of features, we can achieve performance similar or better than using *all* features, our objective will be satisfied. Note that for lower dimensional data, for lower percentage we may not get any feature. In order assess the quality of features we train the system using 70% of the data and test on 30% of the data. This 70-30 partition is repeated 10 times and we report the accuracy's mean ± standard deviation over repeated trials. Note that, for each repeat, all classifier uses the same 70-30 partition to make the result consistent.

The primary objective of this analysis is to assess how good is the Group-lasso based Siamese network in finding useful features and to what extent the number of features can be reduced while sustaining or improving upon 100% baseline accuracy. Unless otherwise noted, comparisons across classifiers on the same dataset are presented as secondary observations and are raised only when the disparity is noteworthy.

TABLE 4: PERFORMANCE OF THE NEAREST NEIGHBOUR CLASSIFIER WITH DIFFERENT PERCENTAGE OF FEATURES CONSIDERING BOTH EUCLIDEAN DISTANCE AND JACCARD SIMILARITY

| | | Percentage of features used | | | | | | | | | | |
|---|---|---|---|---|---|---|---|---|---|---|---|---|
| Dataset | Dist metric | 5% | 10% | 15 | 20 | 25 | 30 | 35 | 40 | 45 | 50 | 100 |
| Flea (6) | EU | | | | | | | 1.00 ± 0.014 | 1.00 ± 0.014 | 1.00 ± 0.014 | 1.00 ± 0.014 | 1.00 ± 0.014 |
| | JS | | | | | | | 0.98 ± 0.030 | 0.98 ± 0.030 | 0.98 ± 0.030 | 1.00 ± 0.014 | 1.00 ± 0.014 |
| Seeds (7) | EU | | | | | | 0.94 ± 0.030 | 0.94 ± 0.030 | 0.94 ± 0.030 | 0.93 ± 0.011 | 0.93 ± 0.011 | 0.92 ± 0.026 |
| | JS | | | | | | 0.94 ± 0.030 | 0.94 ± 0.030 | 0.94 ± 0.030 | 0.93 ± 0.011 | 0.93 ± 0.011 | 0.92 ± 0.026 |
| Pima (8) | EU | | | | | 0.64 ± 0.034 | 0.64 ± 0.034 | 0.64 ± 0.034 | 0.64 ± 0.027 | 0.64 ± 0.027 | 0.71 ± 0.026 | 0.70 ± 0.020 |
| | JS | | | | | 0.67 ± 0.026 | 0.67 ± 0.026 | 0.67 ± 0.026 | 0.70 ± 0.020 | 0.70 ± 0.020 | 0.69 ± 0.018 | 0.70 ± 0.020 |
| Wine (13) | EU | | | | 0.83 ± 0.036 | 0.93 ± 0.018 | 0.93 ± 0.018 | 0.94 ± 0.032 | 0.96 ± 0.017 | 0.96 ± 0.017 | 0.97 ± 0.017 | 0.95 ± 0.018 |
| | JS | | | | 0.88 ± 0.046 | 0.88 ± 0.026 | 0.88 ± 0.026 | 0.94 ± 0.032 | 0.96 ± 0.017 | 0.96 ± 0.017 | 0.97 ± 0.017 | 0.95 ± 0.018 |
| Vehicle (18) | EU | | | 0.45 ± 0.038 | 0.56 ± 0.016 | 0.62 ± 0.020 | 0.69 ± 0.020 | 0.69 ± 0.018 | 0.69 ± 0.020 | 0.68 ± 0.015 | 0.68 ± 0.023 | 0.70 ± 0.022 |
| | JS | | | 0.46 ± 0.031 | 0.60 ± 0.020 | 0.66 ± 0.029 | 0.66 ± 0.027 | 0.68 ± 0.025 | 0.69 ± 0.018 | 0.69 ± 0.020 | 0.69 ± 0.021 | 0.70 ± 0.022 |
| Iono-sphere (33) | EU | | 0.88 ± 0.040 | 0.91 ± 0.021 | 0.90 ± 0.024 | 0.89 ± 0.027 | 0.88 ± 0.020 | 0.90 ± 0.024 | 0.88 ± 0.023 | 0.87 ± 0.023 | 0.87 ± 0.031 | 0.86 ± 0.033 |
| | JS | | 0.80 ± 0.039 | 0.89 ± 0.026 | 0.89 ± 0.025 | 0.89 ± 0.024 | 0.90 ± 0.022 | 0.88 ± 0.030 | 0.88 ± 0.027 | 0.89 ± 0.027 | 0.89 ± 0.023 | 0.86 ± 0.033 |
| Sonar (60) | EU | 0.62 ± 0.025 | 0.77 ± 0.045 | 0.83 ± 0.028 | 0.84 ± 0.037 | 0.86 ± 0.028 | 0.86 ± 0.032 | 0.83 ± 0.038 | 0.84 ± 0.036 | 0.85 ± 0.036 | 0.86 ± 0.028 | 0.85 ± 0.027 |
| | JS | 0.65 ± 0.042 | 0.77 ± 0.044 | 0.79 ± 0.042 | 0.83 ± 0.036 | 0.85 ± 0.021 | 0.88 ± 0.020 | 0.87 ± 0.022 | 0.85 ± 0.036 | 0.85 ± 0.038 | 0.85 ± 0.036 | 0.85 ± 0.027 |
| ORL (1024) | EU | 0.83 ± 0.019 | 0.87 ± 0.022 | 0.89 ± 0.021 | 0.89 ± 0.023 | 0.90 ± 0.023 | 0.91 ± 0.022 | 0.91 ± 0.022 | 0.92 ± 0.019 | 0.92 ± 0.017 | 0.92 ± 0.017 | 0.92 ± 0.022 |
| | JS | 0.79 ± 0.026 | 0.87 ± 0.022 | 0.87 ± 0.019 | 0.88 ± 0.018 | 0.88 ± 0.023 | 0.89 ± 0.016 | 0.90 ± 0.019 | 0.91 ± 0.020 | 0.91 ± 0.018 | 0.92 ± 0.015 | 0.92 ± 0.022 |
| Warp-PIE10P (2420) | EU | 0.99 ± 0.010 | 1.00 ± 0.005 | 0.99 ± 0.008 | 0.99 ± 0.007 | 0.99 ± 0.008 | 0.99 ± 0.008 | 0.99 ± 0.008 | 1.00 ± 0.006 | 1.00 ± 0.006 | 1.00 ± 0.006 | 0.99 ± 0.011 |
| | JS | 0.99 ± 0.010 | 0.99 ± 0.008 | 0.99 ± 0.010 | 0.99 ± 0.010 | 0.99 ± 0.007 | 0.99 ± 0.008 | 0.99 ± 0.008 | 0.99 ± 0.008 | 0.99 ± 0.008 | 0.99 ± 0.008 | 0.99 ± 0.011 |
| Lung (3312) | EU | 0.93 ± 0.025 | 0.95 ± 0.016 | 0.95 ± 0.021 | 0.94 ± 0.022 | 0.95 ± 0.017 | 0.95 ± 0.019 | 0.95 ± 0.020 | 0.95 ± 0.020 | 0.95 ± 0.019 | 0.94 ± 0.021 | 0.95 ± 0.022 |
| | JS | 0.95 ± 0.017 | 0.95 ± 0.021 | 0.95 ± 0.018 | 0.95 ± 0.016 | 0.96 ± 0.015 | 0.96 ± 0.017 | 0.96 ± 0.015 | 0.96 ± 0.017 | 0.96 ± 0.017 | 0.96 ± 0.017 | 0.95 ± 0.022 |
| Gisette (5000) | EU | 0.95 ± 0.005 | 0.92 ± 0.005 | 0.90 ± 0.007 | 0.91 ± 0.006 | 0.91 ± 0.004 | 0.91 ± 0.005 | 0.91 ± 0.008 | 0.91 ± 0.007 | 0.91 ± 0.010 | 0.92 ± 0.006 | 0.93 ± 0.004 |
| | JS | 0.91 ± 0.007 | 0.91 ± 0.005 | 0.91 ± 0.004 | 0.91 ± 0.003 | 0.92 ± 0.006 | 0.92 ± 0.004 | 0.92 ± 0.002 | 0.92 ± 0.006 | 0.92 ± 0.006 | 0.93 ± 0.008 | 0.93 ± 0.004 |
| Leukae-mia (7070) | EU | 0.90 ± 0.061 | 0.89 ± 0.062 | 0.88 ± 0.058 | 0.88 ± 0.041 | 0.89 ± 0.051 | 0.86 ± 0.043 | 0.86 ± 0.041 | 0.85 ± 0.041 | 0.85 ± 0.040 | 0.85 ± 0.041 | 0.87 ± 0.057 |
| | JS | 0.98 ± 0.022 | 0.97 ± 0.022 | 0.95 ± 0.049 | 0.95 ± 0.054 | 0.95 ± 0.066 | 0.95 ± 0.066 | 0.95 ± 0.067 | 0.93 ± 0.062 | 0.91 ± 0.066 | 0.93 ± 0.065 | 0.87 ± 0.057 |

The 1-NN classifier achieves high accuracy using only a small subset of features (Table 4). EUFS and JSFS perform similarly on low- and medium-dimensional datasets, whereas JSFS, as expected, is superior on high-dimensional datasets. The largest improvement is observed on Leukaemia, where JSFS achieves 98% using only 5% of the features (~354), substantially exceeding the 87% full-feature baseline, while EUFS peaks at 90% and declines to 87%. This indicates that JSFS improves both dimensionality reduction and generalization on sparse biomedical data. WarpPIE10P reaches 99–100% accuracy with only 5–10% features, requiring just ~120 of the 2,420 features. For Gisette, EUFS achieves 95% with only 5% features, outperforming both the 93% full-feature baseline and JSFS (91%), although both converge to 93% with all features.

TABLE 5. PERFORMANCE OF THE LINEAR SVM CLASSIFIER WITH DIFFERENT PERCENTAGE OF FEATURES CONSIDERING BOTH EUCLIDEAN DISTANCE AND JACCARD SIMILARITY

| Percentage of features used | | | | | | | | | | | | |
|---|---|---|---|---|---|---|---|---|---|---|---|---|
| Dataset | Dist metric | 5 | 10 | 15 | 20 | 25 | 30 | 35 | 40 | 45 | 50 | 100 |
| Flea (6) | EU | | | | | | | 0.97 ± 0.030 | 0.97 ± 0.030 | 0.97 ± 0.030 | 0.99 ± 0.021 | 1.00 ± 0.000 |
| | JS | | | | | | | 1.00 ± 0.014 | 1.00 ± 0.014 | 1.00 ± 0.014 | 1.00 ± 0.014 | 1.00 ± 0.000 |
| Seeds (7) | EU | | | | | | 0.95 ± 0.019 | 0.95 ± 0.019 | 0.95 ± 0.019 | 0.94 ± 0.028 | 0.94 ± 0.028 | 0.92 ± 0.035 |
| | JS | | | | | | 0.95 ± 0.019 | 0.95 ± 0.019 | 0.95 ± 0.019 | 0.94 ± 0.028 | 0.94 ± 0.028 | 0.92 ± 0.035 |
| Pima (8) | EU | | | | | 0.67 ± 0.026 | 0.67 ± 0.026 | 0.67 ± 0.026 | 0.69 ± 0.014 | 0.69 ± 0.014 | 0.76 ± 0.021 | 0.77 ± 0.017 |
| | JS | | | | | 0.74 ± 0.017 | 0.74 ± 0.017 | 0.74 ± 0.017 | 0.75 ± 0.018 | 0.75 ± 0.018 | 0.75 ± 0.020 | 0.77 ± 0.017 |
| Wine (13) | EU | | | | 0.86 ± 0.038 | 0.94 ± 0.034 | 0.94 ± 0.034 | 0.96 ± 0.020 | 0.97 ± 0.025 | 0.97 ± 0.025 | 0.98 ± 0.017 | 0.96 ± 0.012 |
| | JS | | | | 0.87 ± 0.035 | 0.91 ± 0.027 | 0.91 ± 0.027 | 0.96 ± 0.020 | 0.97 ± 0.025 | 0.97 ± 0.025 | 0.98 ± 0.017 | 0.96 ± 0.012 |
| Vehicle (18) | EU | | | 0.37 ± 0.017 | 0.52 ± 0.025 | 0.57 ± 0.021 | 0.64 ± 0.016 | 0.65 ± 0.024 | 0.72 ± 0.025 | 0.76 ± 0.017 | 0.78 ± 0.013 | 0.80 ± 0.026 |
| | JS | | | 0.38 ± 0.029 | 0.49 ± 0.022 | 0.64 ± 0.021 | 0.65 ± 0.021 | 0.68 ± 0.023 | 0.70 ± 0.023 | 0.72 ± 0.020 | 0.72 ± 0.019 | 0.80 ± 0.026 |
| Iono-sphere (33) | EU | | 0.81 ± 0.029 | 0.81 ± 0.025 | 0.82 ± 0.021 | 0.82 ± 0.023 | 0.81 ± 0.022 | 0.82 ± 0.024 | 0.83 ± 0.037 | 0.82 ± 0.030 | 0.81 ± 0.029 | 0.88 ± 0.025 |
| | JS | | 0.76 ± 0.017 | 0.87 ± 0.020 | 0.87 ± 0.023 | 0.88 ± 0.023 | 0.88 ± 0.020 | 0.88 ± 0.020 | 0.87 ± 0.025 | 0.87 ± 0.023 | 0.89 ± 0.018 | 0.88 ± 0.025 |
| Sonar (60) | EU | 0.53 ± 0.029 | 0.62 ± 0.061 | 0.66 ± 0.045 | 0.78 ± 0.046 | 0.75 ± 0.035 | 0.78 ± 0.040 | 0.78 ± 0.023 | 0.76 ± 0.020 | 0.77 ± 0.019 | 0.76 ± 0.018 | 0.74 ± 0.033 |
| | JS | 0.72 ± 0.038 | 0.69 ± 0.068 | 0.72 ± 0.050 | 0.77 ± 0.043 | 0.76 ± 0.044 | 0.74 ± 0.070 | 0.75 ± 0.039 | 0.76 ± 0.031 | 0.75 ± 0.026 | 0.77 ± 0.032 | 0.74 ± 0.033 |
| ORL (1024) | EU | 0.86 ± 0.018 | 0.92 ± 0.025 | 0.94 ± 0.025 | 0.94 ± 0.021 | 0.95 ± 0.019 | 0.95 ± 0.021 | 0.95 ± 0.019 | 0.95 ± 0.017 | 0.96 ± 0.016 | 0.96 ± 0.017 | 0.96 ± 0.015 |
| | JS | 0.81 ± 0.033 | 0.89 ± 0.017 | 0.91 ± 0.025 | 0.92 ± 0.016 | 0.93 ± 0.018 | 0.94 ± 0.018 | 0.94 ± 0.019 | 0.94 ± 0.015 | 0.95 ± 0.018 | 0.95 ± 0.016 | 0.96 ± 0.015 |
| Warp-PIE10P (2420) | EU | 0.98 ± 0.010 | 0.99 ± 0.011 | 0.98 ± 0.012 | 1.00 ± 0.007 | 0.99 ± 0.010 | 0.99 ± 0.008 | 1.00 ± 0.007 | 1.00 ± 0.007 | 1.00 ± 0.007 | 1.00 ± 0.006 | 1.00 ± 0.007 |
| | JS | 0.98 ± 0.020 | 0.98 ± 0.016 | 0.98 ± 0.016 | 0.98 ± 0.015 | 0.98 ± 0.010 | 0.98 ± 0.010 | 0.98 ± 0.010 | 0.98 ± 0.010 | 0.99 ± 0.009 | 0.99 ± 0.008 | 1.00 ± 0.007 |
| Lung (3312) | EU | 0.97 ± 0.020 | 0.96 ± 0.015 | 0.96 ± 0.017 | 0.96 ± 0.017 | 0.96 ± 0.015 | 0.96 ± 0.014 | 0.96 ± 0.017 | 0.96 ± 0.017 | 0.96 ± 0.015 | 0.96 ± 0.015 | 0.95 ± 0.019 |
| | JS | 0.96 ± 0.020 | 0.97 ± 0.014 | 0.97 ± 0.012 | 0.97 ± 0.011 | 0.97 ± 0.013 | 0.97 ± 0.013 | 0.97 ± 0.012 | 0.97 ± 0.015 | 0.97 ± 0.015 | 0.97 ± 0.018 | 0.95 ± 0.019 |
| Gisette (5000) | EU | 0.98 ± 0.004 | 0.98 ± 0.002 | 0.98 ± 0.002 | 0.99 ± 0.002 | 0.99 ± 0.003 | 0.98 ± 0.002 | 0.98 ± 0.002 | 0.98 ± 0.003 | 0.98 ± 0.003 | 0.98 ± 0.002 | 0.97 ± 0.003 |
| | JS | 0.96 ± 0.004 | 0.96 ± 0.002 | 0.97 ± 0.002 | 0.97 ± 0.002 | 0.97 ± 0.002 | 0.97 ± 0.003 | 0.97 ± 0.002 | 0.97 ± 0.002 | 0.98 ± 0.002 | 0.98 ± 0.002 | 0.97 ± 0.003 |
| Leukae-mia (7070) | EU | 0.95 ± 0.032 | 0.95 ± 0.045 | 0.96 ± 0.045 | 0.96 ± 0.053 | 0.96 ± 0.053 | 0.97 ± 0.041 | 0.97 ± 0.041 | 0.96 ± 0.040 | 0.96 ± 0.040 | 0.96 ± 0.052 | 0.95 ± 0.052 |
| | JS | 0.99 ± 0.018 | 0.99 ± 0.021 | 0.98 ± 0.022 | 0.99 ± 0.021 | 0.98 ± 0.022 | 0.98 ± 0.023 | 0.98 ± 0.023 | 0.98 ± 0.023 | 0.98 ± 0.023 | 0.97 ± 0.022 | 0.95 ± 0.052 |

Table 5 summarizes the performance of the Linear SVM classifier across all datasets. Table 5 reveals that Linear SVM delivers consistently robust and stable performance. On the Seeds dataset, both methods perform equally well, whereas JSFS generally outperforms EUFS on Flea, Pima, Ionosphere, and Leukaemia. In particular for the Leukaemia data set, JSFS achieves 99% accuracy using only 5–10% of the features, outperforming both EUFS (95–96%) and the full-feature baseline (95%). On the other hand, EUFS performs better on the dense image datasets (ORL, WarpPIE10P, and Gisette), with a 1–5% advantage on ORL and perfect accuracy on WarpPIE10P when using 35% or more features; however, JSFS also achieves 98% accuracy on WarpPIE10P using only 5% of the features.

The detailed performance analysis of RBF SVM is reported in Table S1, in the Supplementary materials.  From Table S1, we observe that RBF SVM closely mirrors the stable nature of the Linear SVM, with JSFS achieving its maximum gap on Leukaemia at 5% features (99% for JSFS vs 86% for EUFS). Random Forest exhibits stable performance across most datasets. The detailed results (Table S2) are summarized in the Supplementary Materials. Since Random Forest classifier is not dependent on distances, it works well with features selected by both EUFS and JSFS.

Decision Tree achieves lower accuracies compared to the other classifiers (Table S3, Supplementary Materials), especially on high-dimensional datasets. In the case of Ionosphere, EUFS starts at 88% accuracy with 5% features and remains stable, while JSFS starts lower at 80% but catches up by 10%, while the performance of Sonar for both methods is below the baseline at low percentages, only approaching it at higher percentages of features. It appears that Decision Tree cannot exploit high-dimensional image structures reliably, while Gisette is an exception, with both variants stable at 93%–94% across all percentages.

AdaBoost (Table S4, Supplementary Materials) performs well on low-dimensional datasets. For Pima, JSFS consistently 4% ahead (74% vs. 70%) across all percentages except 100% where both reach 76% accuracy. On ORL, the accuracy is near random performance for all percentages including 100% features; while for WarpPIE10P (41–61%), under both feature selection methods.

TABLE 6. PERFORMANCE OF THE NEURAL NETWORK CLASSIFIER WITH DIFFERENT PERCENTAGE OF FEATURES CONSIDERING BOTH EUCLIDEAN DISTANCE AND JACCARD SIMILARITY

| | | Percentage of features used | | | | | | | | | | |
|---|---|---|---|---|---|---|---|---|---|---|---|---|
| Dataset | Dist metric | 5 | 10 | 15 | 20 | 25 | 30 | 35 | 40 | 45 | 50 | 100 |
| Flea (6) | EU | | | | | | | 0.97 ± 0.042 | 0.98 ± 0.042 | 0.97 ± 0.042 | 1.00 ± 0.014 | 1.00 ± 0.000 |
| | JS | | | | | | | 1.00 ± 0.000 | 1.00 ± 0.000 | 1.00 ± 0.000 | 1.00 ± 0.014 | 1.00 ± 0.000 |
| Seeds (7) | EU | | | | | | 0.96 ± 0.018 | 0.96 ± 0.019 | 0.96 ± 0.019 | 0.94 ± 0.032 | 0.94 ± 0.033 | 0.93 ± 0.033 |
| | JS | | | | | | 0.96 ± 0.019 | 0.96 ± 0.018 | 0.96 ± 0.019 | 0.95 ± 0.031 | 0.94 ± 0.032 | 0.94 ± 0.030 |
| Pima (8) | EU | | | | | 0.70 ± 0.024 | 0.70 ± 0.029 | 0.70 ± 0.028 | 0.69 ± 0.029 | 0.70 ± 0.030 | 0.76 ± 0.017 | 0.74 ± 0.024 |
| | JS | | | | | 0.75 ± 0.025 | 0.75 ± 0.024 | 0.75 ± 0.022 | 0.74 ± 0.023 | 0.75 ± 0.024 | 0.74 ± 0.024 | 0.75 ± 0.019 |
| Wine (13) | EU | | | | 0.88 ± 0.042 | 0.95 ± 0.028 | 0.95 ± 0.028 | 0.96 ± 0.023 | 0.98 ± 0.023 | 0.98 ± 0.017 | 0.98 ± 0.013 | 0.98 ± 0.015 |
| | JS | | | | 0.87 ± 0.043 | 0.90 ± 0.029 | 0.89 ± 0.030 | 0.96 ± 0.025 | 0.98 ± 0.017 | 0.98 ± 0.020 | 0.98 ± 0.017 | 0.98 ± 0.017 |
| Vehicle (18) | EU | | | 0.52 ± 0.036 | 0.62 ± 0.024 | 0.71 ± 0.024 | 0.77 ± 0.020 | 0.78 ± 0.029 | 0.79 ± 0.025 | 0.80 ± 0.026 | 0.81 ± 0.019 | 0.84 ± 0.025 |
| | JS | | | 0.51 ± 0.021 | 0.62 ± 0.020 | 0.70 ± 0.017 | 0.72 ± 0.021 | 0.75 ± 0.012 | 0.77 ± 0.019 | 0.79 ± 0.026 | 0.78 ± 0.026 | 0.84 ± 0.018 |
| Iono-sphere (33) | EU | | 0.90 ± 0.014 | 0.93 ± 0.017 | 0.92 ± 0.022 | 0.92 ± 0.020 | 0.92 ± 0.022 | 0.94 ± 0.026 | 0.93 ± 0.030 | 0.93 ± 0.026 | 0.92 ± 0.027 | 0.92 ± 0.031 |
| | JS | | 0.84 ± 0.028 | 0.92 ± 0.021 | 0.92 ± 0.031 | 0.92 ± 0.016 | 0.93 ± 0.019 | 0.93 ± 0.022 | 0.94 ± 0.018 | 0.94 ± 0.018 | 0.93 ± 0.018 | 0.92 ± 0.028 |
| Sonar (60) | EU | 0.65 ± 0.039 | 0.75 ± 0.048 | 0.83 ± 0.025 | 0.87 ± 0.034 | 0.85 ± 0.040 | 0.86 ± 0.036 | 0.87 ± 0.039 | 0.87 ± 0.038 | 0.87 ± 0.025 | 0.85 ± 0.035 | 0.83 ± 0.038 |
| | JS | 0.72 ± 0.040 | 0.81 ± 0.047 | 0.80 ± 0.047 | 0.87 ± 0.039 | 0.84 ± 0.038 | 0.87 ± 0.048 | 0.86 ± 0.033 | 0.86 ± 0.033 | 0.87 ± 0.032 | 0.86 ± 0.030 | 0.84 ± 0.031 |
| ORL (1024) | EU | 0.83 ± 0.034 | 0.87 ± 0.018 | 0.91 ± 0.013 | 0.91 ± 0.022 | 0.92 ± 0.020 | 0.93 ± 0.018 | 0.94 ± 0.013 | 0.93 ± 0.016 | 0.93 ± 0.023 | 0.93 ± 0.024 | 0.93 ± 0.013 |
| | JS | 0.77 ± 0.029 | 0.86 ± 0.016 | 0.86 ± 0.028 | 0.87 ± 0.035 | 0.90 ± 0.021 | 0.90 ± 0.024 | 0.90 ± 0.022 | 0.90 ± 0.025 | 0.91 ± 0.017 | 0.92 ± 0.024 | 0.93 ± 0.022 |
| Warp-PIE10P (2420) | EU | 1.00 ± 0.007 | 1.00 ± 0.005 | 0.99 ± 0.008 | 0.99 ± 0.011 | 0.99 ± 0.008 | 1.00 ± 0.007 | 1.00 ± 0.005 | 0.99 ± 0.011 | 1.00 ± 0.006 | 0.99 ± 0.011 | 0.99 ± 0.015 |
| | JS | 0.99 ± 0.010 | 0.99 ± 0.012 | 0.98 ± 0.017 | 0.99 ± 0.009 | 0.99 ± 0.010 | 0.99 ± 0.013 | 1.00 ± 0.007 | 0.99 ± 0.011 | 1.00 ± 0.006 | 0.99 ± 0.011 | 0.99 ± 0.014 |
| Lung (3312) | EU | 0.95 ± 0.013 | 0.96 ± 0.017 | 0.96 ± 0.017 | 0.95 ± 0.014 | 0.95 ± 0.017 | 0.95 ± 0.026 | 0.94 ± 0.019 | 0.96 ± 0.025 | 0.95 ± 0.019 | 0.95 ± 0.014 | 0.91 ± 0.033 |
| | JS | 0.97 ± 0.023 | 0.97 ± 0.014 | 0.96 ± 0.019 | 0.97 ± 0.019 | 0.96 ± 0.012 | 0.96 ± 0.022 | 0.96 ± 0.019 | 0.95 ± 0.014 | 0.95 ± 0.020 | 0.95 ± 0.014 | 0.93 ± 0.026 |
| Gisette (5000) | EU | 0.99 ± 0.002 | 0.99 ± 0.001 | 0.99 ± 0.003 | 0.99 ± 0.002 | 0.99 ± 0.002 | 0.98 ± 0.002 | 0.98 ± 0.002 | 0.98 ± 0.003 | 0.98 ± 0.002 | 0.98 ± 0.003 | 0.97 ± 0.002 |
| | JS | 0.97 ± 0.003 | 0.98 ± 0.002 | 0.98 ± 0.002 | 0.98 ± 0.002 | 0.98 ± 0.002 | 0.98 ± 0.002 | 0.98 ± 0.002 | 0.98 ± 0.003 | 0.98 ± 0.002 | 0.98 ± 0.002 | 0.98 ± 0.004 |
| Leukae-mia (7070) | EU | 0.89 ± 0.051 | 0.87 ± 0.077 | 0.90 ± 0.101 | 0.90 ± 0.069 | 0.86 ± 0.072 | 0.90 ± 0.061 | 0.85 ± 0.074 | 0.88 ± 0.071 | 0.87 ± 0.078 | 0.88 ± 0.061 | 0.89 ± 0.068 |
| | JS | 0.96 ± 0.052 | 0.94 ± 0.051 | 0.95 ± 0.038 | 0.92 ± 0.083 | 0.94 ± 0.062 | 0.97 ± 0.030 | 0.92 ± 0.050 | 0.93 ± 0.037 | 0.90 ± 0.047 | 0.90 ± 0.073 | 0.87 ± 0.052 |

For MLP (Table 6), JSFS achieves perfect accuracy on Flea with only 5% features, whereas EUFS reaches 100% at 20%. Both methods slightly exceed the full-feature baseline on Seeds and perform similarly on Vehicle. On high-dimensional datasets, EUFS achieves perfect accuracy on WarpPIE10P with just 5% features (~121 of 2,420), representing one of the strongest results of the study. EUFS also outperforms JSFS on ORL at lower feature percentages, although the gap narrows as more features are included. On Lung, JSFS maintains a consistent 1–2% advantage over EUFS, with both exceeding the full-feature baseline. Overall, MLP combined with JSFS on Leukaemia and with EUFS on WarpPIE10P delivers the strongest individual performances.

Naïve Bayes performs consistently across datasets (Table S5, Supplementary Materials). JSFS generally provides higher accuracy on Pima and Leukaemia, particularly with 5–10% features, while EUFS has a slight advantage on ORL and WarpPIE10P. Most datasets achieve performance close to the full-feature baseline using only 10–30% of the features, demonstrating the effectiveness of both feature selection approaches.

**Do the proposed systems produce quality embeddings also?**

As discussed earlier, our objective was to select useful features for classification without sacrificing the quality of embeddings that the Siamese network should produce. We have already demonstrated that both variants of the system select very useful features for classification. The effectiveness of the selected features has been established using eight widely different types of classifiers. Here we want to assess whether the selected features also produce faithful embeddings of the data satisfying the primary goal of the Siamese Network. In order to demonstrate this, we have considered two datasets, one high dimensional (Gisette) dataset and one moderate dimensional (Ionosphere) dataset. We have projected each data sets as well as their embeddings to 2-dimension by manifold learning algorithms UMAP and t-SNE. This results in six figures for each data set.

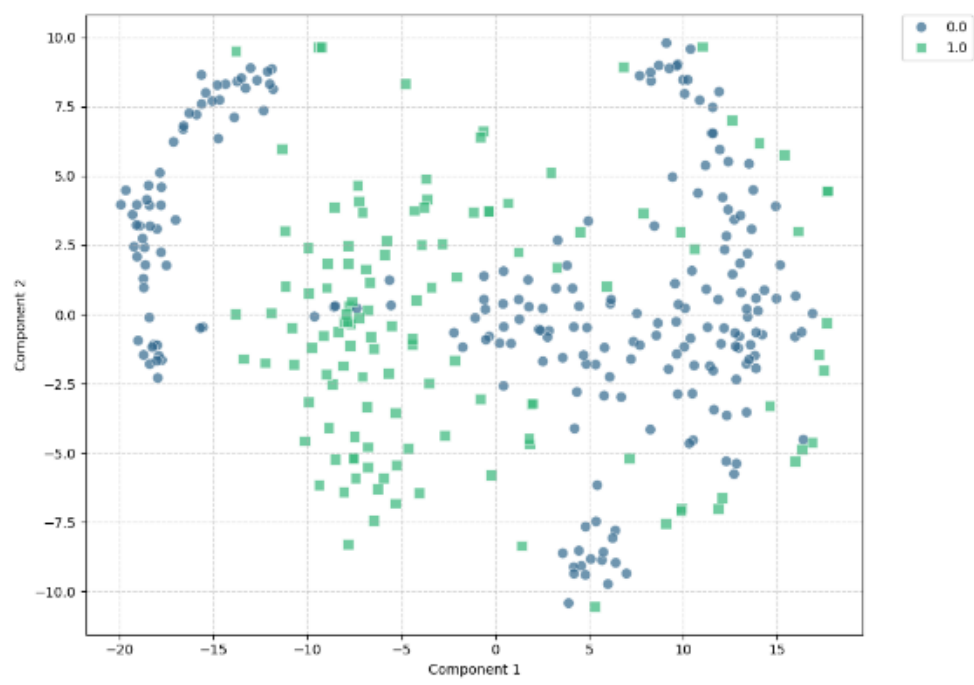

Fig. 2a. Projection of original data by t-SNE

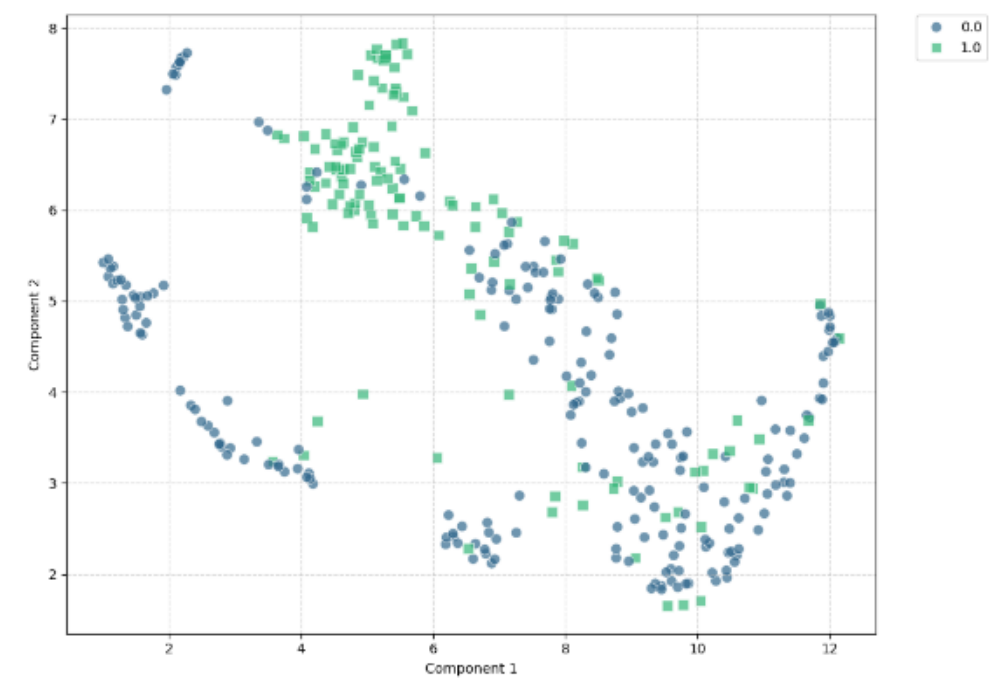

Fig. 2b. Projection of original data by UMAP

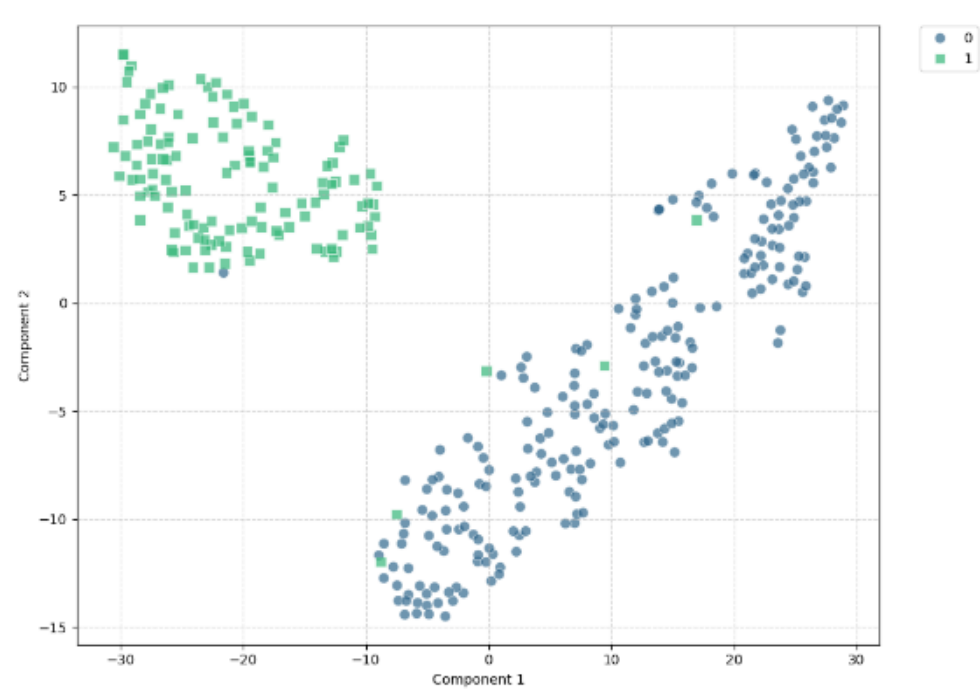

Fig. 2c. Jaccard Embeddings by t-SNE

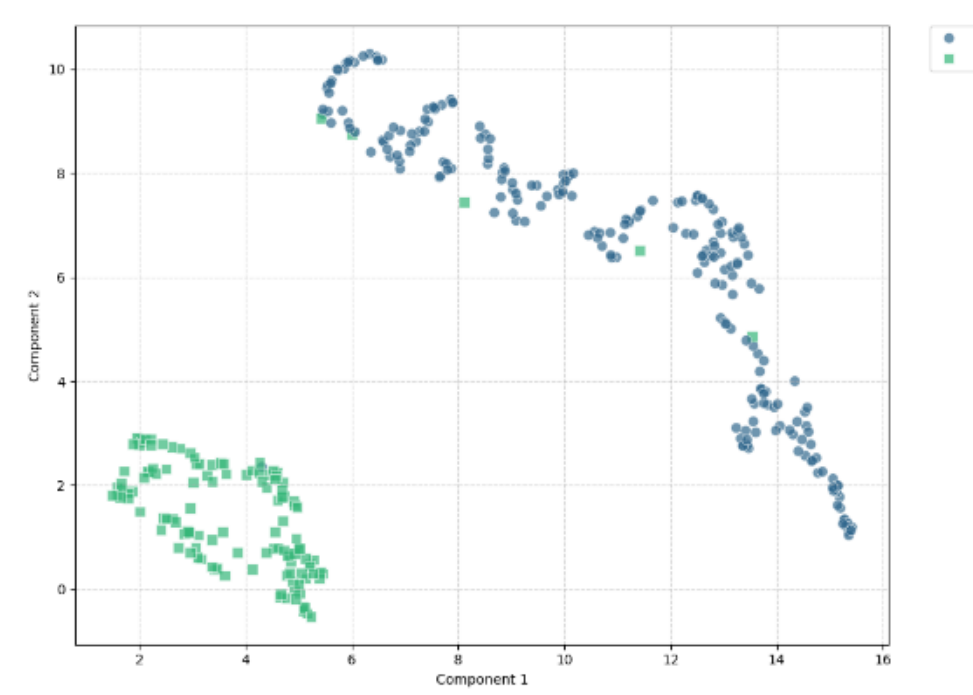

Fig. 2d. Jaccard Embeddings by UMAP

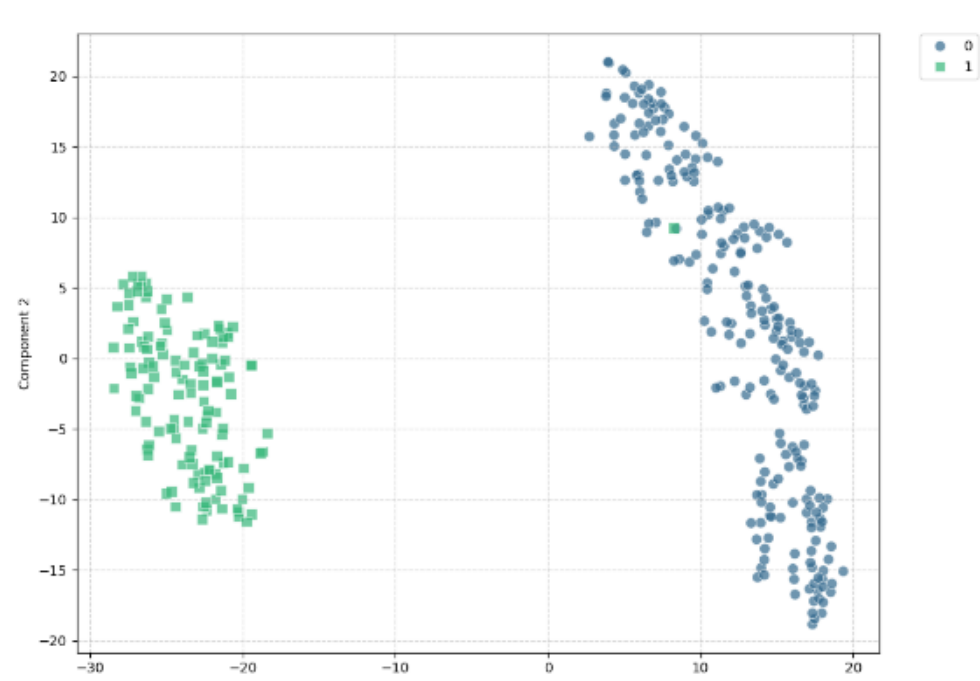

Fig. 2e. Euclidean Embeddings by t-SNE

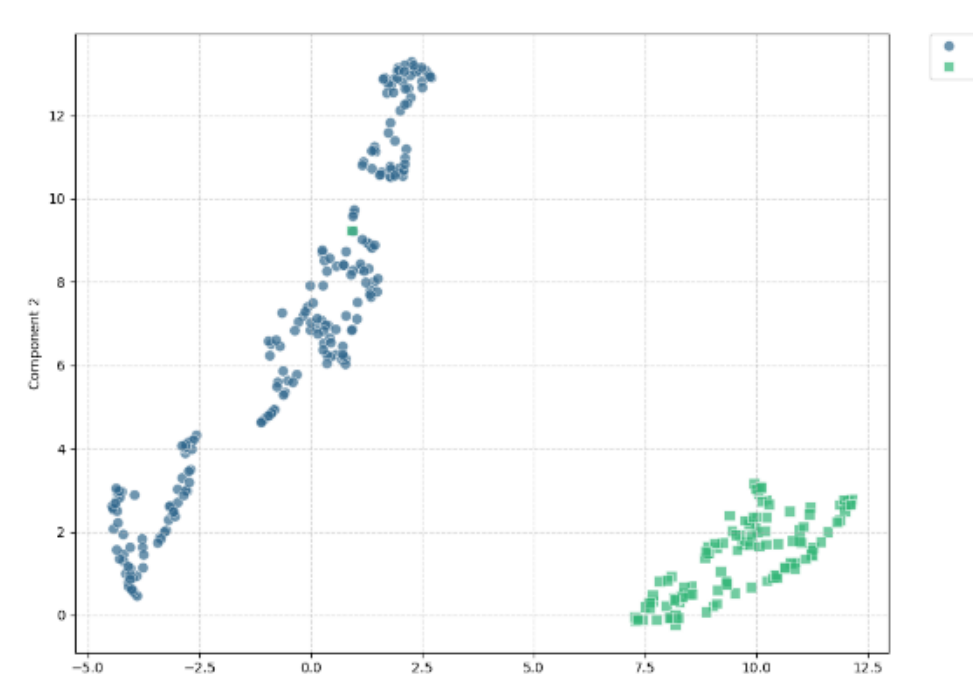

Fig. 2f. Euclidean Embeddings by UMAP

Figure 2. UMAP and t-SNE plots of 2-dimensional projections of the original Ionosphere data, embeddings produced by Jaccard Similarity-based network and the embeddings produced by Euclidean distance-based networks.

Fig. 2a and Fig. 2b depict the t-SNE and UMAP plots of the original Ionosphere dataset, respectively. These figures reveal that classes are not very well separated. Fig. 2c and Fig. 2d display, respectively, the t-SNE and UMAP plots of the embeddings of Ionosphere dataset generated by the Jaccard Similarity based network. These figures show that classes are well separated in the 2-dimensional projected space. The t-SNE and UMAP plots of the embeddings of the same dataset generated by the Euclidean distance-based network are shown in Fig. 2e and Fig. 2f, respectively. These figures also reveal that classes are well separated in the embedded space.

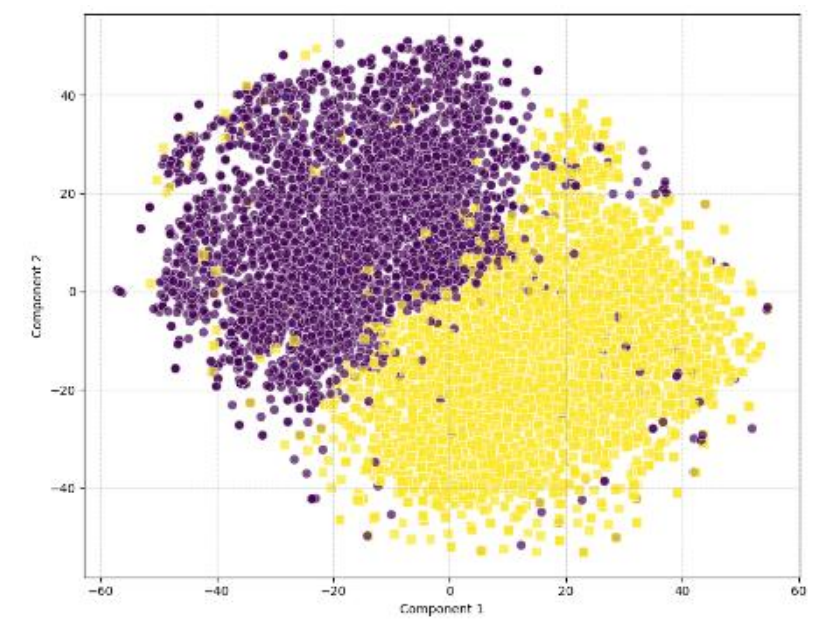

Fig. 3a. Projection of original data by t-SNE

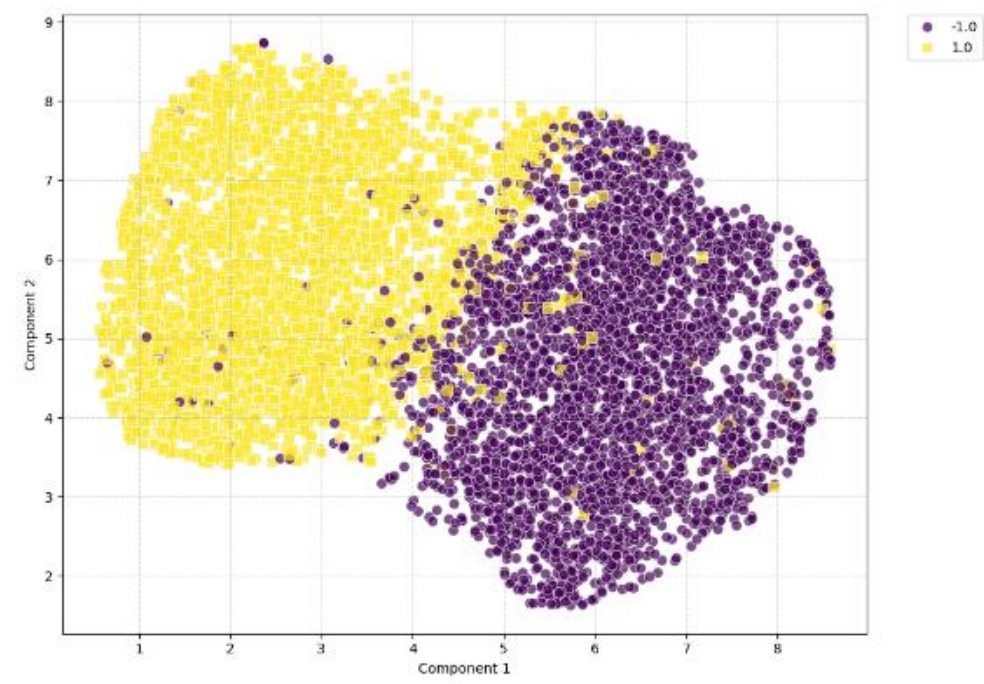

Fig. 3b. Projection of original data by UMAP

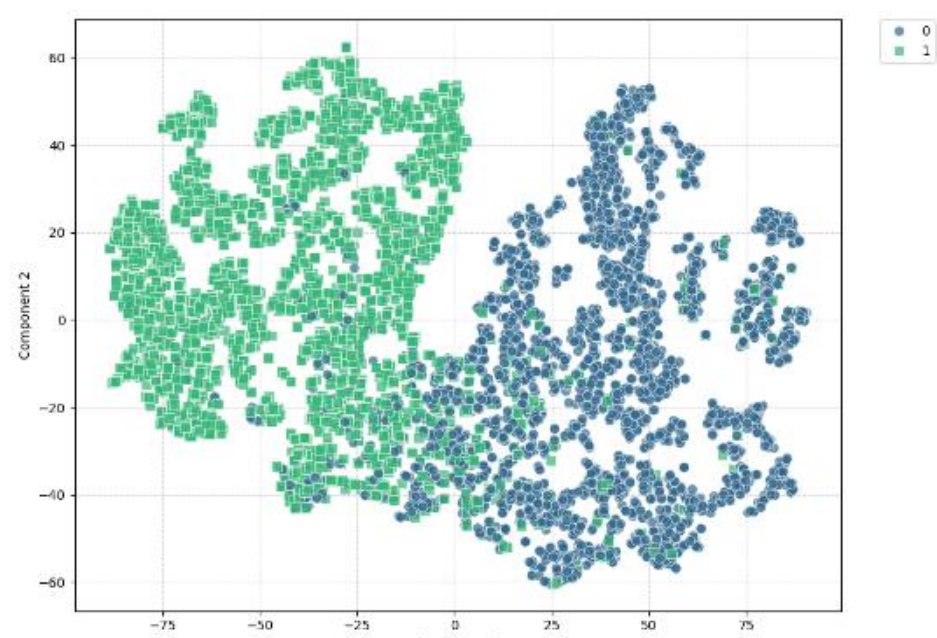

Fig. 3c. Jaccard Embeddings by t-SNE

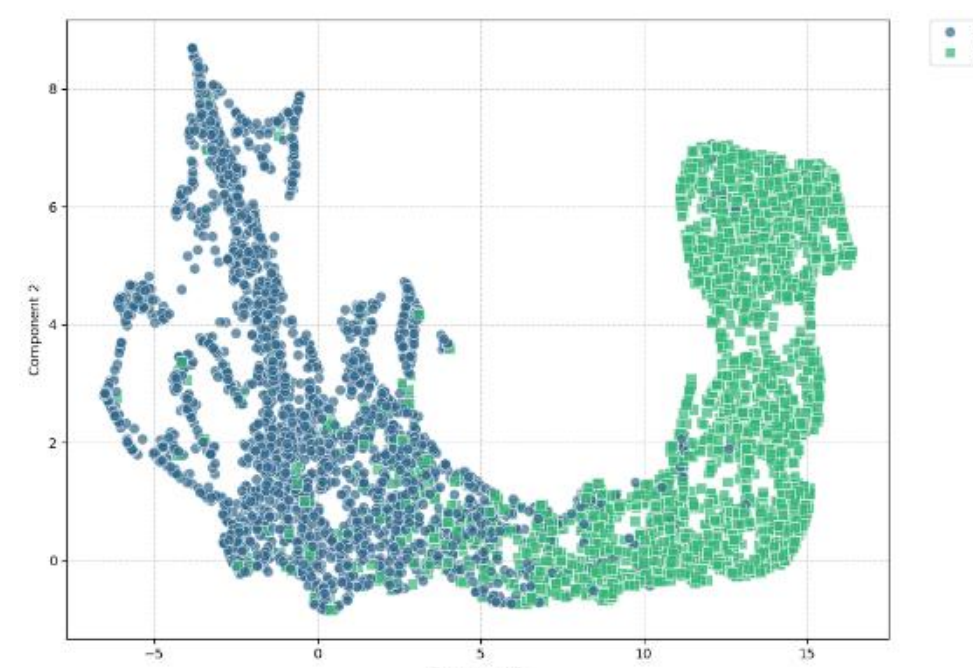

Fig. 3d. Jaccard Embeddings by UMAP

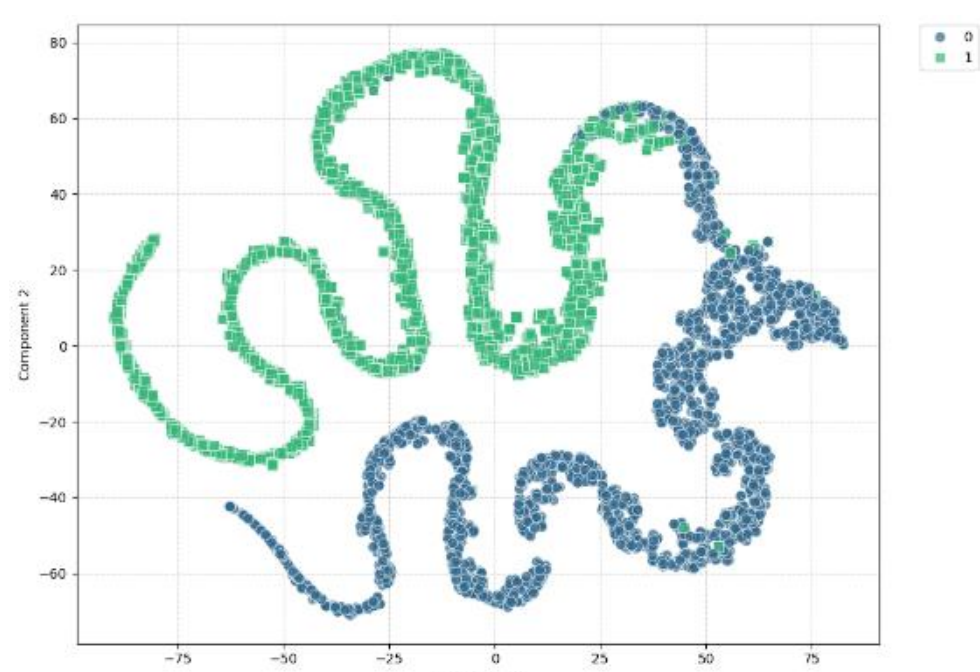

Fig. 3e. Euclidean Embeddings by t-SNE

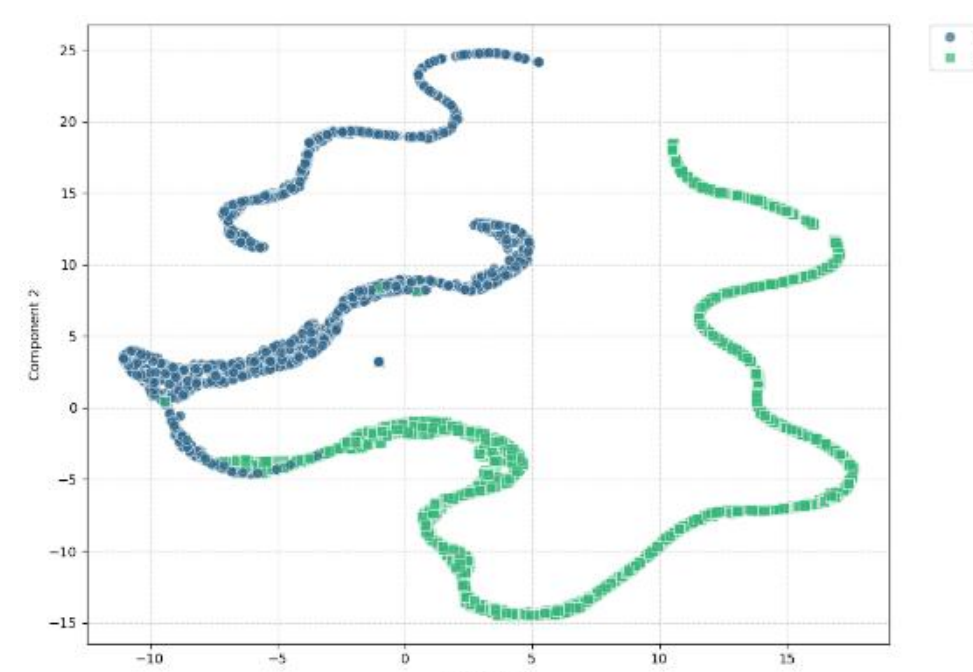

Fig. 3f. Euclidean Embeddings by UMAP

Figure 3. UMAP and t-SNE plots of 2-dimensional projections of the original Gisette data, embeddings produced by Jaccard Similarity-based network and the embeddings produced by Euclidean distance-based networks.

Next, we consider the Gisette data set which is in 5000-dimension. Fig. 3a and Fig. 3b depict the t-SNE and UMAP plot of the original dataset, respectively. From these figures, we observe that the classes are not separated; t-SNE shows a significant overlap between the classes. Fig. 3c and Fig. 3d depict the t-SNE and UMAP projections of the embeddings of the Gisette generated by the Jaccard based similarity. These two figures reveal that classes are well separated except for a few points. Similarly, for the embeddings produced by the Euclidean distance-based network, Fig. 3e and Fig. 3f, generated by t-SNE and UMAP, respectively, again confirm that the generated embeddings are well separated expect for few points. It is worth noting that for the Euclidean-distance based embeddings the 2-D projections form a few curve segments, while the Jaccard Similarity-based embeddings produce compact clusters with a small overlap in the 2-D projections. This is expected and desirable because Gissete dataset consists of handwritten digits 4 and 9. Thus, most of the instances of 4 and 9 are expected to form compact clusters in the input space with some overlap when there is confusion between 9 and 4. As we explained earlier, this may be viewed as an indicator that for high-dimensional data, Jaccard similarity-based network is better than the Enclidean distance-based network.

## VI. Discussion

For the low-dimensional datasets (Flea, Seeds, Pima and Wine), the maximum accuracy is generally achieved using a small subset of features. Flea consistently reaches 97–100% accuracy with only 20–35% features, while Seeds performs almost similarly under both EUFS and JSFS across all classifiers. The Pima data set benefits more with lower feature percentages from JSFS - improves accuracy by about 3–7% over EUFS.

For the medium-dimensional datasets (Vehicle, Ionosphere and Sonar), accuracy generally increases with additional features before stabilizing at around 20–40% features. On the other hand, Ionosphere and Sonar achieve near-maximum performance with only 10–20% features, with RBF-SVMs and MLPs outperforming tree-based classifiers. However, the differences between EUFS and JSFS are small.

For high-dimensional datasets the difference in performance of the features selected by the two networks are generally more substantial. For ORL and WarpPIE10P, EUFS generally performs better than JSFS with Linear SVM, RBF SVM and MLPs. WarpPIE10P reaches 98–100% accuracy with only 5–10% features, while Gisette achieves 97–99% accuracy using the same proportion. This suggests that only a small subset of informative features is required. In contrast, Leukaemia strongly favors JSFS, achieving 98–99% accuracy with only 5% features compared with 86–95% for EUFS. In several cases, 5–10% selected features outperform the full-feature baseline, indicating improved generalization after removing redundant and irrelevant features.

Among the classifiers, Linear SVM, RBF SVM and Neural Networks consistently deliver the highest and most consistent accuracies, often reaching the peak performance with only a small fraction of the features. Random Forest also performs robustly with minimal differences between EUFS and JSFS.

Decision Tree and AdaBoost perform comparatively worse, particularly on high-dimensional datasets. Decision Tree exhibits lower accuracy, especially on ORL and WarpPIE10P as compared to low dimensional datasets, similarly AdaBoost performs well on low-dimensional datasets but fails to generalize on complex image datasets, achieving only 8–13% accuracy on ORL and 41–61% on WarpPIE10P.

Overall, EUFS and JSFS exhibit complementary strengths. EUFS is generally more effective for dense image datasets, whereas JSFS performs better on sparse, high-dimensional datasets, particularly on Leukaemia. Overall, the proposed feature selection framework reduces the feature space by 70–95% while maintaining or improving classification accuracy across datasets with diverse characteristics.

## VII. Conclusion

We have proposed a mechanism for selection of useful features using the Siamese network. In this regard, we have used a group-lasso based regularizer on the connection weights between the input and the first hidden layer along with the usual contrastive loss function. Apparently, such a system will select features to achieve the goal of Siamese network. We hypothesised that the selected features should also be good discriminator for classification. The rationale behind such a hypothesis was that if the features can produce embeddings to bring points from the same class closer and to push points from different classes apart, it has to make use of discriminating power of the features. We have demonstrated using extensive experiments that our hypothesis was true. In this context, we have developed two variants of the network; one using Euclidean distance and the other using a Jaccard similarity based on distance measure. The effectiveness of the method is demonstrated using a set of 12 data sets from different domains and having widely varying dimensions. As explained and also demonstrated that for high-dimensional data set, Siamese network with Jaccard Similarity based distance measure works better. Further, we have demonstrated that the feature-selection Siamese networks can also generate faithful embeddings of the data satisfying the primary objective of the Siamese network.

Like any other study, ours is not free from limitations. Using cross-validation, we can choose the most desirable architecture as well as the penalty parameter, $\lambda$, given adequate computing power. We have not explored the effectiveness of the triplet loss. Another interesting extension would be to augment the loss function with a third term that considers the discriminating power of the embedding. This may require adding a third tower augmented with some layers of MLP.